# Survival Prediction of Heart Failure Patients using Stacked Ensemble Machine Learning Algorithm

S.M Mehedi Zaman[1], Wasay Mahmood Qureshi[2], Md. Mohsin Sarker Raihan[3], Ocean Monjur[4] and Abdullah Bin Shams[5]*

Department of Electrical and Electronic Engineering, Islamic University of Technology, Gazipur 1704, Bangladesh[1,2]
Department of Biomedical Engineering Khulna University of Engineering & Technology, Khulna 9203, Bangladesh[3]
Department of Computer Science and Engineering, Islamic University of Technology, Gazipur 1704, Bangladesh[4]
Department of Electrical & Computer Engineering, University of Toronto, Toronto, Ontario M5S 3G4, Canada[5]
email: mehedizaman@iut-dhaka.edu[1], wasaymahmood@iut-dhaka.edu[2], msr.raihan@gmail.com[3],
o.monjur@gmail.com[4] and abdullahbinshams@gmail.com[5*]

*Corresponding author

*Abstract*— cardiovascular disease, especially heart failure is one of the major health hazard issues of our time and is a leading cause of death worldwide. Advancement in data mining techniques using machine learning (ML) models is paving promising prediction approaches. Data mining is the process of converting massive volumes of raw data created by the healthcare institutions into meaningful information that can aid in making predictions and crucial decisions. Collecting various follow-up data from patients who have had heart failures, analyzing those data, and utilizing several ML models to predict the survival possibility of cardiovascular patients is the key aim of this study. Due to the imbalance of the classes in the dataset, Synthetic Minority Oversampling Technique (SMOTE) has been implemented. Two unsupervised models (K-Means and Fuzzy C-Means clustering) and three supervised classifiers (Random Forest, XGBoost and Decision Tree) have been used in our study. After thorough investigation, our results demonstrate a superior performance of the supervised ML algorithms over unsupervised models. Moreover, we designed and propose a supervised stacked ensemble learning model that can achieve an accuracy, precision, recall and F1 score of 99.98%. Our study shows that only certain attributes collected from the patients are imperative to successfully predict the surviving possibility post heart failure, using supervised ML algorithms.

## I. INTRODUCTION

One of the most common forms of heart disease is heart failure and it implies to the gradual weakening of heart muscles to pump blood to our bodies. At present, the number of heart failure cases is approximately 64.34 million worldwide and is rising rapidly over the years [1]. However, not all the patients having heart failures die because of it though the mortality figures are worrisome. There are certain attributes which can, if utilized properly, predict whether the patient will survive or not. These attributes include serum sodium, ejection fraction, serum creatinine, blood-pressure, age etc. [2]. Generally, people who have had heart failures are observed in hospital for several days after the incident. Several kinds of parameters are recorded in a regular basis from the patients' blood to further monitor their health condition. Some data are recorded outside the hematological parameters such as: Age, sex, smoking status etc. The challenge then comes to analyze the recorded data and to find correlations among them to predict whether the patients' condition is deteriorating or getting better. Machine learning (ML) algorithms process user data to learn and predict outcomes automatically. They have the ability to adapt to different scenarios and pass judgements based on previous learnings [3]. These algorithm models are being used extensively to healthcare sectors especially in multiple disease diagnosis [4]. Analyzing the follow-up data, post heart failure, from the patients is significant in terms of mortality prediction. Recent methods of data analysis include various ML models such as: Logistic Regression, Support Vector Machine (SVM), Random Forest, Naïve Bayes, etc. [5]. While these models have certain features to predict quite accurately from variety of data, they need proper training to be able to be implemented in an unknown set of data based on practical aspects. Nevertheless, the machine learning approaches have been rather successful in comparison with previous reports and as a matter of fact, the prediction accuracies and precisions are improving every year [2, 5, 6]. Çağatay Berke Erdaş and Didem Ölçer (2020) used multiple algorithms and various feature selections to get the best possible outcomes from the dataset collected from the follow-up data of heart failure patients. Their results showed accuracies of 86% for 1Rule, and 84% for Random Forest and SVM algorithms [19]. Another study by Davide Chicco & Giuseppe Jurman (2020) with the same dataset, reported comparatively lower accuracies of 73.70% with Decision Tree, 74% with Random Forest and 83.30% with Logistic Regression algorithms [5]. Abid Ishaq et al., (2021) demonstrated a significant improvement in the prediction accuracy of 92.62% using Extra Trees Classifier (ETC) [20]. The studies conducted by Jaymin Patel et al., (2016) and Fahd Saleh Alotaibi (2019), with a different dataset, focused on 13 features. The latter acquired a high accuracy score of 93.19% whereas the former demonstrated a moderate accuracy of 83.40% [21, 6]. This paper focuses towards an improvement strategy in the prediction metrics (accuracy, precision, recall & F1) of survival rates of heart failure patients utilizing the dataset provided by UCI and the case study of Ahmad et al., (2017). In our study, both unsupervised and supervised ML algorithms were implemented to assess which category of ML model is suitable for predicting the survuval possibility. In unsupervised learning, K-Means & Fuzzy C-Means Clustering methods and in supervised learning Decision Tree, Random Forrest & XGBoost algorithms are used. Based on the best performing classifiers, using 12 features, we design and propose a supervised stacked ensemble learning method. Our results, show a significant improvement in the prediction metrics of survival chances. The structure of our ensemble model consists of three base learners and one meta learner. The proposed model can also be implemented in real-life practical scenarios, as it is an autonomous method that is capable of being used in an emergency. The analysis itself is instantaneous and thus can be administered in crucial moments.

## II. DATA FAMILIARIZATION

The dataset contains various features of around 300 patients who suffered from left ventricular systolic dysfunction. The 12 columns consist of several attributes of these patients following their heart failures. The follow-up period averages to 130 days [2]. Table 1 provides a quick overview of the attributes. Our target predictor is the "Survival Event" attribute which has a categorical value meaning 1 = "Passed away" and 0 = "Survived".

**TABLE 1: Description of the dataset used in this study.**

| Attribute | Description |
| --- | --- |
| Age | Patient age (Years) |
| Anemia | Lack of healthy red blood cells (Categorical) |
| Creatinine Phosphokinase | Creatinine Phosphokinase enzyme in the blood (mcg/L) |
| Diabetes | Diabetic status of patient (Categorical) |
| Ejection Fraction | Portion of blood leaving the heart in each cardiac contraction (%) |
| High blood pressure | Status of blood pressure (Categorical) |
| Platelets | Blood platelet count (kiloplatelets/mL) |
| Serum Creatinine | Blood creatinine level (mg/dL) |
| Serum Sodium | Blood sodium level (mEq/L) |
| Sex | Gender of patient (Categorical) |
| Smoking | Status of smoking (Categorical) |
| Time | Follow-up period (Days) |
| Survival Event | Incident of a surviving patient during the follow-up period (Categorical) |

## III. APPROACH AND METHODOLOGY

Fig. 1 graphically illustrates the chronological steps used in this study for the analysis of the data for survival prediction.

### A. Dataset collection

The dataset was collected from the case study of Ahmad et al., (2017) which is available from the University of California Irvine Machine Learning Repository [2].

### B. Data pre-processing

We started by checking for any missing values in the dataset. Fortunately, there were none. However, in the "Age" and "Platelets" columns, a few float values were found and were rounded off to integer values as the majority of the values were whole numbers.

### C. SMOTE analysis

The target attribute "Survival Event", had an initial imbalance in the dataset as the mentioned column had uneven events of survival (203) and death (96) [8]. This class distribution is skewed and training a ML algorithm with such dataset results in a biasness towards the majority class. This reduces the predictive performance of the ML model, especially for the minority class.

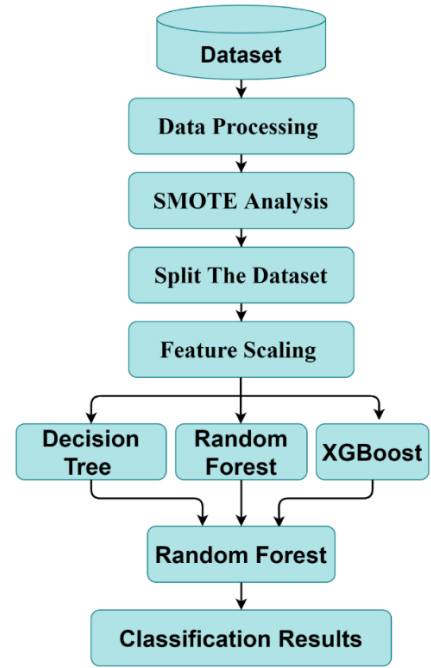

Fig. 1. Methodology flowchart.

Synthetic Minority Oversampling Technique (SMOTE) algorithm was applied to oversample the dataset to mitigate the imbalance problem [9]. As we can see from Fig. 2, after oversampling, we had equal number (203) of events. Next, we leaned into the feature scaling section where data normalization is performed [10].

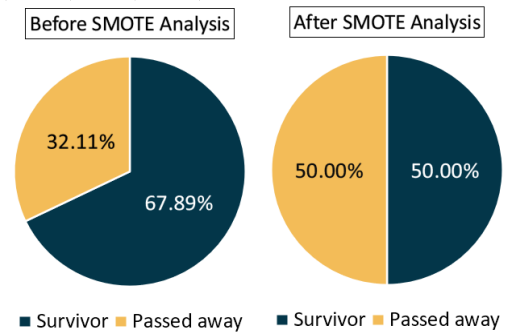

Fig. 2. Percentage of survivor and passed away before and after SMOTE

### D. Splitting the dataset

The training and testing data were split into 80% and 20% respectively as our machine learning algorithms will first train themselves with the training dataset to predict the outcome and then they will be able to test their prediction metrics on the test dataset [11].

*E. Feature scaling*

Numerous machine learning estimators in the dataset require standardization; if the individual features do not closely resemble standard normally distributed data, they may perform poorly: Gaussian with a mean of zero (0) and unit variance. In practice, we frequently ignore the distribution's shape and simply convert the data to centre it by deleting each feature's mean value, then scale it by dividing non-constant characteristics by their standard deviation [12].

*F. Applied Machine Learning Algorithms*

In this portion of our study, we have implemented both unsupervised and supervised learning. Unsupervised learning, in which a machine accepts inputs but does not receive any target outputs which are supervised or rewards from the overall domain, is another kind of machine learning classification. It may seem a little strange to speculate on what the machine itself might be able to learn in the event that it's not getting any response from its surroundings. On the contrary, a framework for unsupervised learning is possible to be developed on the basis that the machine's aim is to create decision-making representations of the input, forecasting follow-up inputs, and communicating inputs to another machine efficiently, among other things [23]. Data clustering is a significant approach to unsupervised learning. It is a technique for determining the cluster structure of a data set based on the degree of similarity between clusters and the degree of dissimilarity between clusters [24]. This clustering can be assessed through resemblance in different metrics such as distance. Clustering is utilized to point out subgroups within heterogeneous data, such that each cluster has a higher degree of homogeneity than the whole [25]. In this paper, we have implemented two types of clustering. They are as follows:

*1) K-means clustering*

This clustering method groups data in such a way that each piece of data can only belong to one cluster. Here, assigning each point to one of the initial clusters, each cluster center is replaced by the cluster's mean point. The process is repeatedly done until finally it reaches a convergence [26]. Though it is meant for big data analysis, it can be used in relatively smaller datasets as well. One of the key aspects of k-means clustering algorithm is to set the proper value of "k". The dataset we are dealing with has two target variables and hence we already know the best clustering value of "k" which is 2.

*2) Fuzzy C-means clustering*

This is an overlapping technique where clustering data are of fuzzy types. The conventional perception of probability is extended through the fuzziness. Assigning each point a different degree of membership, this clustering allows a point's membership in multiple clusters to be shared. Thus, resulting the concept of fuzzy boundaries, as opposed to the more traditional concept of unambiguous boundaries [26]. Moreover, fuzzy clustering is not capable of pointing out the exact number of clusters although it worked quite good in our dataset.

For the supervised learnings, three machine learning algorithms were used as base learners. Then, using the predicted outcomes of these base learners, a meta learner algorithm (Random Forest for this study) was administered to get higher accuracy. What the meta learner does is that, it takes the predicted values of the base learners as inputs and splits them into a hold-out set and an implementable set. The implementable set is used as the first layer of the stacked learning and trained accordingly. After that, the predictions made in the first layer is used into the hold-out set to make certain that the predictions are novel. Next, these predictions are used as training sets for the meta learner which happens to be the best performer among the base learners in our case [7]. The three base learner machine learning algorithms are as follows:

*3) Decision tree (DT)*

Decision Tree method is one of the most popular learning algorithms that belongs to the family of supervised learning algorithms. This refers to a tree-like model where all the internal nodes (including the root node) symbolize attribute tests and the leaf nodes represent the outcome of the test. It is explicitly used as a tool for decision making [13]. Decision Trees are a successive model that efficiently and cohesively unifies a series of basic tests in which a numeric feature is compared to a threshold value in each test. This method is used for handling regression and classification problems. To formulate an efficient decision tree, it is critical to control the size of the decision tree. These measures are based on theories of information entropy, such as information gain, gain ratio, and distance-based measure. Information gain is a term that refers to a change in entropy H from one state to another. The term "entropy" refers to

$$H(T) = -\sum_{y \in Y} P(Y) \log_b P(Y)$$

where T denotes a collection of labeled training instances, y denotes an instance label, and P(y) denotes the probability of drawing an instance with label y from T. Gaining information is defined as

$$IG(T, T_{left}, T_{right}) = H_T - \left(\frac{n_{left}}{n_{total}}\right) \times H(T_{left}) - \left(\frac{n_{right}}{n_{total}}\right) \times H(T_{right})$$

Here $T_{left}$ and $T_{right}$ are the subsets of $T$ created by a decision rule. $n_{total}$, $n_{left}$ and $n_{right}$ refer to the number of examples in the respective sets [25].

Controlling the size requires avoiding overfitting during the learning process. In decision tree-based machine learning techniques, pruning has been shown to be the most effective method of dealing with overfitting [24]. There are several popular decision tree algorithms based on the target variables such as: Iterative Dichotomies 3 (ID3), Successor of ID3 (C4.5), Classification and Regression Tree (CART) [14]. The majority of decision tree learning algorithms are derived from a fundamental algorithm that performs a top-down greedy search approach in the space of decision trees [24]. The C4.5 algorithm is the successor to the ID3 algorithm, which employs a dimension pruning rule. It offers a greedy search method for a decision tree that never returns to re-evaluate previous alternatives. This algorithm has difficulties training non-valued attributes in attributes, but it is also non-incremental and inexpensive.

*4) Random forest (RF)*

Random Forest, one of the most powerful ensemble methods in Machine learning, uses a top-down approach to find relevant features, comparing them to the set of original attributes [15]. This is a technique that is frequently used in Classification and Regression problems. A random forest algorithm is considered as a collection of decision trees. Three components comprise a decision tree: decision nodes, leaf nodes, and a root node. Additionally, this algorithm is an extension of the decision tree algorithm that incorporates the bagging concepts of the Boosting methodology. It generates a 'forest' that is trained using bagging or bootstrap aggregation. Bagging is a meta-algorithm used in ensembles learning to increase the accuracy of algorithms. The original random forest model selects models using all decision tree classifiers and incorporates all decision trees into the random forest model for voting [16]. The single decision tree is extremely sensitive to changes in the data. It is prone to overfit to data noise. The Random Forest with a single tree will also overfit to data, as it is identical to a single decision tree. The greater the number of trees it has, the more accurate the result, and the less likely it is to overfit. Sometimes to avoid overfitting in this technique, the primary goal is to optimize a tuning parameter that controls the number of randomly chosen features used to grow each tree from the data that were bootstrapped. The random forest algorithm performs the overall estimation and has the advantage of feature selection automatically. Overall estimation will be carried out by the random forest algorithm, which has the advantage of automatic feature selection [27]. The random forest processes two types of data: random attribute data and original data. The training set accounts for approximately two-thirds of the total data set, with the remainder referred to as out-of-bag (oob) data and used to estimate unbiased classification error as new trees are added to the forest [15]. This method integrates the simplicity of decision trees with the flexibility of Decisions trees, resulting in a significant increase in accuracy. We chose Random Forest because it outperforms all other existing MLAs, including decision trees and neural networks, in terms of predictive accuracy.

*5) XGBoost (XGB)*

XGBoost (Extreme Gradient Boost) has been a competitive tool among Machine Learning methods due to its features such as multithreading parallel computation and high prediction accuracy. This is an efficient and end-to-end scalable implementation of the Gradient Boosting Machine (GBM) to achieve state-of-the-art results [17]. This algorithm, belonging to the family of Supervised learning methods, refers to the process of inferring a predictive model from a collection of labeled training examples. This predictive model can then be used to predict new unobserved instances. XGBoost is a generalised gradient boosting implementation that incorporates a regularisation term for anti-overfitting and support for arbitrary differentiable loss functions [25]. Rather than optimizing for simple squared error loss, a two-part objective function is defined: a loss function over the training set and a regularisation term penalizing the model's complexity:

$$Obj = \sum_i L(x_i, \hat{x}_i) + \sum_k \Omega(f_k)$$

For a given training dataset, difference between predictive value and true label will be measured by the loss function $L(x_i, \hat{x}_i)$. Tree's complexity $f_k$ is denoted by $\Omega(f_k)$ which in the XGBoost algorithm as [26]

$$\Omega(f_k) = \gamma T + \frac{1}{2}\lambda \omega^2$$

This boosting technique performs well when the dataset is unbalanced. Gradient boosting is a technique that involves creating new models that predict the residuals or errors of prior models and then combining them to make the final prediction. Additionally, this is a supervised learning method that employs a technique known as boosting to achieve more accurate models than other algorithms. Extreme Gradient boosting is so named because it employs a gradient descent algorithm to reduce the loss associated with the addition of new models. This approach is applicable to predictive modelling problems involving regression and classification. XGBoost outperforms Gradient Boosting using parallel computing and advanced regularization (L1 & L2).

*G. Performance Evaluation Metrics (PEM)*

To compare the predictive performances of the algorithms, four evaluation metrics are used, which are computed from the confusion matrix.

|  | | Actual Values | |
|---|---|---|---|
|  | | Yes | No |
| Predicted Value | Yes | TP | FP |
|  | No | FN | TN |

Fig. 3. Typical confusion matrix.

In Fig. 3, a typical confusion matrix is depicted, with the abbreviations TP, TN, FP, and FN represents True Positive, True Negative, False Positive, and False Negative, respectively. Now, the evaluation metrics are as follows:

*1) Accuracy*

The accuracy metric, in general, is defined as the proportion of correct predictions to the total number of occurrences examined [18].

$$\text{Accuracy (acc)} = \frac{TP+TN}{TP+FP+TN+FN}$$

*2) Precision*

Precision is a metric for determining the proportion of accurately predicted positive patterns in a positive class relative to the total predicted patterns [18].

$$\text{Precision (p)} = \frac{TP}{TP+FP}$$

*3) Recall*

The percentage of correctly classified positive patterns is measured by recall [18].

$$\text{Recall (r)} = \frac{TP}{TP+TN}$$

*4) F1 score*

This metric represents the harmonic mean of recall and precision values [18].

## IV. RESULTS AND DISCUSSION

The selected features and their respective correlation with the survival possibility post heart failure are illustrated in Fig.4. The correlation value ranges from -1 to +1. The closer the value of a certain predictor is towards +/-1, the better is the correlation of that to the target variable. The "Time" feature had the highest positive correlation with the target "Survival Event". This suggests that the follow-up period post heart failure is highly associated with reducing the mortality risk. Next important positively correlated attribute are the "Ejection Fraction" and "Serum Sodium".

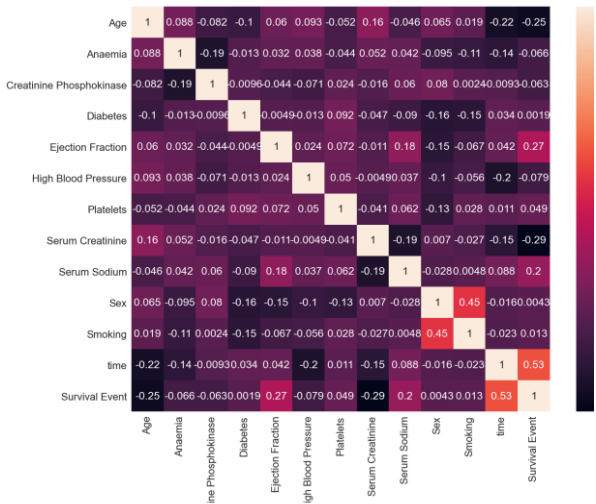

Fig. 4. Feature correlation matrix.

A higher ejection fraction indicates that the heart is pumping enough oxygenated blood around the body, which is very crucial for survival, in general. Heart failure can cause a reduction in circulating blood volume which decreases the blood pressure. As a compensatory response to maintain the blood pressure, sodium is retained in the body [31]. Therefore, a high sodium level in the blood indicates adequate blood circulation which increases the chances of survival. Age and serum creatinine shows strong negative correlations. Ageing and rising mortality risk post heart failure is a common phenomenon, as the heart muscles weakens when atherosclerosis develops with time. This reduces the flow of oxygenated blood supply to the cardiac muscles, and increases the possibility of a heart attack or a stroke reducing the surviving chances post heart failure [32]. High creatinine level is an indicator of insufficient renal blood flow which reflects a reduced cardiac output decreasing the survival probability [33]. The rest of the biomarkers seems to have very minimal significance in predicting the survival chances. We will now discuss the performance of the ML algorithms based on these attributes.

At first, we discuss the performance of unsupervised models to analyse, cluster and classify the unlabelled dataset. This learning method is very popular to find hidden patterns. In our study, we applied the two well-known clustering algorithms, K-Means and Fuzzy C-Means. The models could successfully for two clusters (Survivor/Passed away) as can be seen in Fig.5 & Fig.6 for K-Means Clustering and Fuzzy C-Means Clustering respectively. Despite forming two distinct clusters, the separation is not wide reducing the cluster quality. This limited the clustering accuracy to 62.24% & 52.45% for K-Means & Fuzzy C-Means algorithms respectively. Also, no additional pattern can be observed from the data.

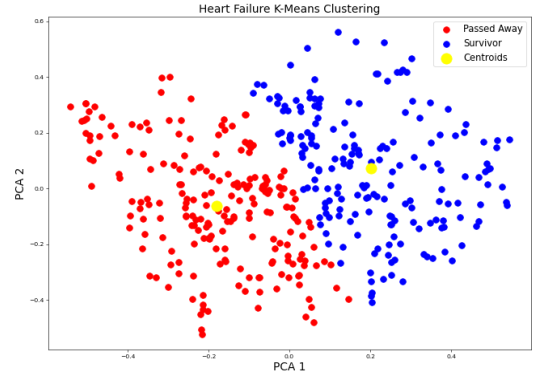

Fig. 5. K-Means Clustering.

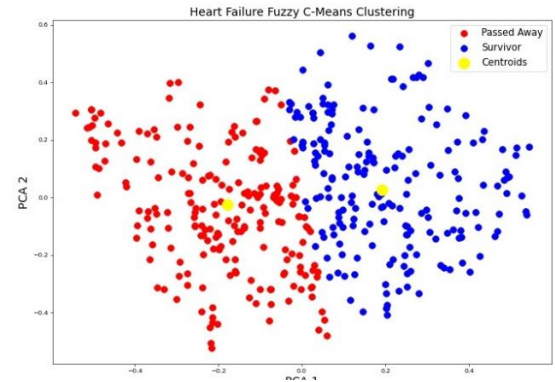

Fig. 6. Fuzzy C-Means Clustering.

For supervised learning, we have implemented several algorithms from which we shortlisted the best three performing ones. They are: Random Forest, XGBoost and Decision Tree and their performance metrics are illustrated in Fig.7. The former two algorithms show similar and better performances than Decision Tree. This performance is closer to the works mentioned earlier. The performance metrics of the prediction can further improve using ensemble learning [4].

The results of our proposed stacked ensemble machine learning algorithm are compared with the abovementioned studies and illustrated in Table 2. It can be noted that, with similar number of features and with the same dataset, our algorithm outperforms the contemporary studies. This can be mainly attributed to our stacked ensemble architecture.

**TABLE 2 : Comparison of our results with previous studies.**

| References | Features | Algorithm | Accuracy | Precision | Recall | F1 Score |
|---|---|---|---|---|---|---|
| Davide Chicco & Giuseppe Jurman (2020) | 2 | RF, LR, DT, SVM, KNN | 83.30% | N/A | N/A | 71.40% |
| Jaymin Patel et al., (2016) | 13 | J48, Logistic Model Tree, RF | 83.40% | N/A | N/A | N/A |
| Çağatay Berke Erdaş and Didem Ölçer (2020) | 13 | 1Rule, RF, SVM, Multi-Layer Perceptron, NB | 86% | 94% | 86% | 90% |
| Abid Ishaq et al., (2021) | 9 | DT, ADB, LR, RF, Extra Trees Classifier, SVM | 92.62% | 93% | 93% | 93% |
| Fahd Saleh Alotaibi (2019) | 13 | DT, NB, RF, LR, SVM | 93.19% | N/A | N/A | N/A |
| **This study** | **12** | **Stacked Ensemble Machine Learning** | **99.98%** | **99.98%** | **99.98%** | **99.98%** |

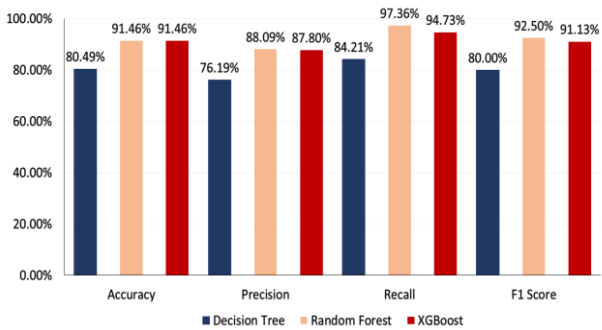

Fig. 7. Performance metrics of base learning model.

To design a stacked ensemble model, we used these three best performing algorithms as the base learner, whose predictions are then input into the meta learner algorithm i.e., Random Forest, to predict the survival possibility. The architecture of our ensemble model can be seen in Fig.1. The whole procedure is done sequentially from base learners to meta learner, and we obtained a significant improvement in our performance to 99.98% for all of the four metrics which are accuracy, precision, recall and F1 score respectively, as shown in Fig.8.

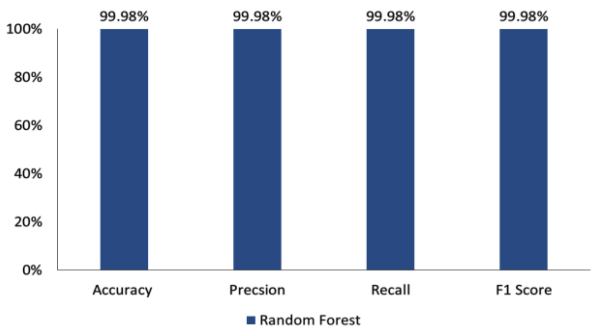

Fig. 8. Performance metrics of stacked ensemble learning model.

Measurement of the overall performance is critical in ML. Hence, we utilize the AUC (Area Under the Curve) metric and ROC (Receiver Operating Characteristics) curve. The AUC - ROC curve is a performance evaluation for classification problems at different thresholds. AUC represents the degree or measure of separability, whereas ROC is a probability curve. It indicates how well the model can distinguish between classes [22]. In our study, as the Random Forest algorithm showed the most promise, in Fig. 6, we plotted the ROC and the AUC for both the base and meta learner of the Random Forest algorithm. The AUC score was found to be 0.968 and 0.99 for base and meta learners respectively.

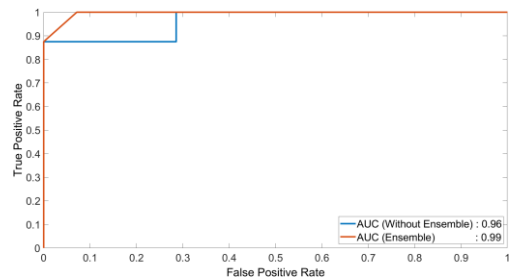

Fig. 9. ROC Curve of base and meta learners of Random Forest algorithm.

## V. CONCLUSION

In this study, we extensively focus in the survival prediction of patients post heart failure. Instead of using the conventional statistical analysis to manually predict survival rates, we proposed and successfully implemented a stacked ensemble machine learning algorithm for real-time autonomous prediction of a patient survival possibility after a heart failure using the follow-up data. We used both supervised and unsupervised learning models. After thorough investigation, our results demonstrate a superior performance of the supervised algorithms over unsupervised models. To further improve the prediction accuracy we designed and proposed an ensemble machine learning algorithm. By integrating three base and one meta learners, we obtained a boost in the performance of 99.98% in accuracy, precision, recall and F1 score, with an AUC score of 0.99. Using similar number of features and with the same dataset, our algorithm outperforms the contemporary studies. This can be mainly attributed to our stacked ensemble architecture. Further, the challenge of the imbalance dataset was mitigated by SMOTE analysis. We plan to implement our algorithm into a mobile application to ease and keep track of the follow-up data of heart failure patients more effectively. As a result, our research can be translated into real life healthcare scenario. As the data analysis method used here for predicting is instantaneous, appropriate measures can be taken for critically ill patients immediately once their test results are available from the laboratory.